\documentclass{article}

\usepackage[preprint]{corl_2026}

\usepackage[utf8]{inputenc}
\usepackage[T1]{fontenc}
\usepackage{graphicx}
\usepackage{booktabs}
\usepackage{amsmath}
\usepackage{xcolor}

\hypersetup{
  pdfauthor={Praveen Selvaraj, Lorenzo Uttini, Ville Kuosmanen},
  pdftitle={Rollout then Recycle: 3,718 Labelled Real-Robot Episodes from Benchmarking 7 Policies on 12 Tasks}
}

\makeatletter
\renewcommand{\@noticestring}{%
  Accepted to the ``Oops, I Erred'' Workshop at CoRL 2026.%
}
\makeatother

\newcommand{\hfds}[1]{\href{https://huggingface.co/datasets/#1}{\nolinkurl{#1}}}

\title{Rollout then Recycle: 3{,}718 Labelled Real-Robot Episodes from Benchmarking 7 Policies on 12 Tasks}

\author{
  Praveen Selvaraj$^{1,*}$ \quad Lorenzo Uttini$^{1,2,*}$ \quad Ville Kuosmanen$^{1,*}$\\[6pt]
  \mdseries $^{1}$Armnet \quad $^{2}$Imperial College London\\[6pt]
  {\small\mdseries $^{*}$Equal contribution}
}

\begin{document}
\maketitle

\begin{abstract}
We release all 3{,}718 episodes from the ArmnetBench~v0.1
evaluation~\citep{armnetbench2026}, which contains 600 teleoperated demonstrations and
3{,}118 rollouts of seven imitation-learning and vision-language-action
policies on 12 tabletop tasks, recorded on low-cost SO-101 arms; 73\% of the
rollouts are failures. Every policy for a task was trained or fine-tuned on the
same 50 demonstrations and rolled out on the same cell, so the corpus contains
many policies' failures on shared tasks. An operator labelled every rollout
\texttt{successful}, \texttt{suboptimal} or \texttt{failure}. Each episode has three camera views,
joint states and actions and is available on the Hugging Face Hub in
LeRobot~v3.0 and a reward-model format. This note documents how the episodes were collected and the limitations that could affect their use.
\end{abstract}


\section{Introduction}

Evaluation rollouts are an obvious source of real-robot failure data. Every
benchmark produces them, often already labelled, as a by-product. And yet large datasets
such as Open~X-Embodiment~\citep{openx2024} and DROID~\citep{khazatsky2024droid}
consist mostly of demonstrations. In a comparison of eight representative
real-world benchmarks~\citep{armnetbench2026}, none releases its rollouts as a
labelled corpus. ArmnetBench~v0.1 evaluated seven policies on 12 tasks and kept
every scored rollout. All policies for a task were trained or fine-tuned on the
same 50 demonstrations and rolled out on the same cell, so the corpus holds the successes and failures of many policies under one setup.
The hardware and leaderboard are described in~\citep{armnetbench2026}.

\section{Data Collection}
\label{sec:collection}

All episodes were recorded during the ArmnetBench~v0.1 evaluation using the
setup, tasks, protocol and labelling scheme described below.

\textbf{Setup.} Episodes were recorded on 3 co-located SO-101
cells~\citep{cadene2024lerobot}, two single-arm and one bimanual, each with
three cameras at 20\,fps. State and action are joint and gripper positions,
6-D on the single-arm cells and 12-D on the bimanual cell.

\begin{figure}[t]
\centering
\includegraphics[width=0.9\textwidth]{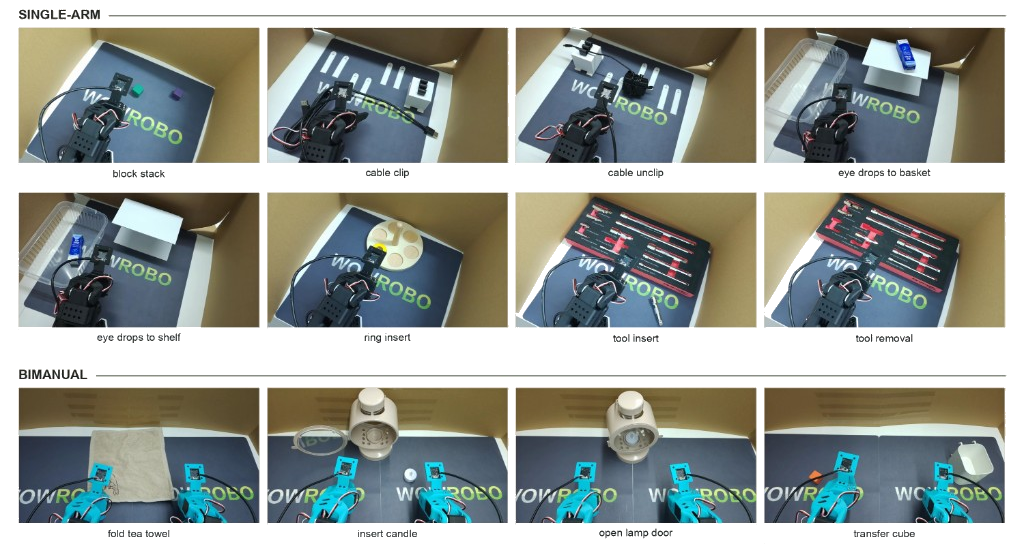}
\caption{Each of the 12 tasks as viewed from the \texttt{top} camera: eight single-arm tasks in the
upper two rows and four bimanual tasks in the bottom row.}
\label{fig:tasks}
\end{figure}

\textbf{Tasks and policies.} Each of the 12 tasks (Figure~\ref{fig:tasks})
has 50 teleoperated demonstrations, successful by construction. ACT~\citep{zhao2023aloha},
Diffusion Policy~\citep{chi2023diffusion}, SmolVLA~\citep{smolvla2025},
$\pi_0$~\citep{black2024pi0}, $\pi_{0.5}$~\citep{pi05_2025},
GR00T~N1.7~\citep{nvidia2025gr00t} and MolmoAct~2~\citep{molmoact2_2026}
were each trained or fine-tuned per task on those demonstrations. All of the checkpoints are publicly available for ease of reproduction.

\textbf{Protocol.} The benchmark targets 30 rollouts per task--policy pair.
Additional sessions repeated MolmoAct~2 on all 12 tasks with an earlier
checkpoint, ACT on four tasks and SmolVLA on five, so pairs have 29--60
rollouts. Before each rollout the operator reset the scene and randomised
object placement by hand. There was no set time limit and the operator stopped each
rollout. One operator supervised all three cells.

\textbf{Labels.} The operator labelled each rollout \texttt{successful} if the
goal was achieved cleanly, \texttt{suboptimal} if it was achieved with a
poor-quality finish and \texttt{failure} if it was not achieved. Successful and suboptimal rollouts were cut at the moment of completion, while failures ran until the operator stopped them. Two rollouts with incorrect resets and any corrupted or unlabelled recordings were excluded.

\section{Dataset Contents and Access}

The 3{,}718 episodes (${\sim}$26 hours) comprise 600 demonstrations and 3{,}118
policy rollouts. The rollouts break down into 724 \texttt{successful} (23.2\%), 106 \texttt{suboptimal}
(3.4\%) and 2{,}288 \texttt{failure} (73.4\%) as detailed in Tables~\ref{tab:composition} and~\ref{tab:tasks}.

\begin{table}[t]
\centering
\begin{minipage}[t]{0.46\textwidth}
\centering
\small
\setlength{\tabcolsep}{3pt}
\caption{Rollouts per policy.}
\label{tab:composition}
\begin{tabular}{@{}lrrrr@{}}
\toprule
\textbf{Source} & \textbf{$n$} & \textbf{Succ.} & \textbf{Sub.} & \textbf{Fail.} \\
\midrule
$\pi_{0.5}$         & 359 & 171 & 14 & 174 \\
$\pi_0$             & 359 & 126 & 19 & 214 \\
GR00T~N1.7          & 360 & 106 & 13 & 241 \\
Diffusion           & 360 &  96 & 11 & 253 \\
ACT                 & 480 &  70 & 14 & 396 \\
MolmoAct~2          & 720 &  92 & 15 & 613 \\
SmolVLA             & 480 &  63 & 20 & 397 \\
\midrule
Rollouts            & 3{,}118 & 724 & 106 & 2{,}288 \\
Demos               & 600 & 600 & 0 & 0 \\
\bottomrule
\end{tabular}
\end{minipage}\hfill
\begin{minipage}[t]{0.5\textwidth}
\centering
\small
\setlength{\tabcolsep}{3pt}
\caption{Rollouts per task.}
\label{tab:tasks}
\begin{tabular}{@{}lrrrr@{}}
\toprule
\textbf{Task} & \textbf{$n$} & \textbf{Succ.} & \textbf{Sub.} & \textbf{Fail.} \\
\midrule
\texttt{block\_stack}           & 255 & 37  & 0  & 218 \\
\texttt{cable\_clip}            & 270 & 0   & 0  & 270 \\
\texttt{cable\_unclip}          & 240 & 93  & 0  & 147 \\
\texttt{eye\_drops\_to\_basket} & 240 & 120 & 0  & 120 \\
\texttt{eye\_drops\_to\_shelf}  & 299 & 79  & 0  & 220 \\
\texttt{ring\_insert}           & 255 & 82  & 0  & 173 \\
\texttt{tool\_insert}           & 300 & 54  & 43 & 203 \\
\texttt{tool\_removal}          & 240 & 50  & 9  & 181 \\
\midrule
\texttt{fold\_tea\_towel}       & 300 & 69  & 39 & 192 \\
\texttt{insert\_candle}         & 240 & 23  & 15 & 202 \\
\texttt{open\_lamp\_door}       & 240 & 54  & 0  & 186 \\
\texttt{transfer\_cube}         & 239 & 63  & 0  & 176 \\
\bottomrule
\end{tabular}
\end{minipage}
\end{table}

The release is a Hugging Face
collection\footnote{\url{https://huggingface.co/collections/armnet/armnetbench-v01}} with two LeRobot~v3.0
repositories, \hfds{armnet/armnetbench_v01_lerobot_so101} with 2{,}499 episodes
and \hfds{armnet/armnetbench_v01_lerobot_bimanual_so101} with 1{,}219. A third
repository, \hfds{armnet/armnetbench_v01_robometer}, holds the same episodes in
the training format of the Robometer reward model~\citep{liang2026robometer},
which has three rows per episode, one per camera, with a 240p clip in each row. In the LeRobot v3.0 format, each episode carries
\texttt{success\_class} and \texttt{policy\_type}, plus the evaluated
checkpoint as \texttt{policy\_repo\_id}. Demonstrations are labelled
\texttt{successful}, with \texttt{policy\_type} set to \texttt{teleoperated}.
Each frame carries \texttt{next.done}, which is true only on the last frame of
an episode. It also carries a sparse terminal reward, \texttt{next.reward},
which is 1.0 at the end of every \texttt{successful} episode (demonstrations
included) and 0 elsewhere; \texttt{suboptimal} episodes therefore receive no
reward. The episodes have also been submitted to
the Oopsie-Data project.\footnote{\url{https://oopsie-data.com/how-contributing/}}

\section{Uses}

The corpus supports methods that require both successful and failed attempts collected under a shared setup. Every task pairs 50 demonstrations with 239--300 labelled rollouts from several policies trained on those demonstrations. These properties support three main uses:
\begin{itemize}
    \item \textbf{Success detection and reward modelling.} The outcome labels
    can supervise success classifiers and episode-level reward models.
    Holding out a policy or task tests generalisation to unseen failure modes.
    \item \textbf{Learning from mixed-quality data.} The labels allow rollout
    data to be filtered for imitation, used to condition policies on outcome
    quality or converted into terminal rewards for offline RL.
    \item \textbf{World modelling.} Failed rollouts show action-conditioned world
    models how unsuccessful interactions evolve rather than exposing them
    only to successful demonstrations, helping to remove their bias towards successful task completions. 
\end{itemize}

\section{Limitations}

The release has limitations arising from its labels, its collection protocol
and its metadata.

\textbf{Labels and episode ends.} Each label reflects a single operator's
judgement. Only 106 rollouts (3.4\%), all on four tasks, are
\texttt{suboptimal}, so the labels are effectively binary and carry no
partial-progress signal. Because successful and suboptimal rollouts were
trimmed at completion while failures ran until the operator stopped them,
episode length and trailing idle motion leak the label; models should be
trained on length-matched or clipped inputs. For the same reason,
\texttt{next.done} on a failure marks an operator stop rather than a terminal
state, which matters for offline RL.

\textbf{Collection protocol.} All data come from three cells in one room with
one cell per task and ambient lighting. Initial object positions were not
recorded. Conditions also varied between policies. Some objects wore down over
the course of collection. Two camera problems affected every policy except
MolmoAct~2~\citep{armnetbench2026}: the \texttt{front} camera on the cable and
eye-drops cell was misaligned and the bimanual \texttt{right\_wrist} view was
slightly blurry. Policy identity may therefore be inferable from the images
alone.

\textbf{Metadata and splits.} The metadata does not mark which 2{,}518
rollouts form the benchmark; only the additional MolmoAct~2 sessions can be
separated, via \texttt{policy\_repo\_id}, so the leaderboard cannot be
reproduced exactly from the release. Finally, the Robometer export stores one
row per camera, so each episode appears three times; splits must be made by
episode rather than by row to keep views of the same rollout out of both
training and test sets.

\clearpage

\bibliography{references}

\end{document}